\documentclass[ijaa,nonblindrev]{informs3_noinforms}
\usepackage{hyperref}
\usepackage{pdfsync, cleveref}
\usepackage{enumerate}
\usepackage{mathrsfs}
\usepackage{multirow,subcaption}
\usepackage{bbm}
\usepackage{makecell}
\usepackage{xspace}
\usepackage[dvipsnames]{xcolor}
\usepackage[most]{tcolorbox} %

\DoubleSpacedXII

\usepackage{algorithm}
\usepackage{algpseudocode}
\usepackage{tikz}
\newcommand{\bI}{\mathbbm{1}}

\newcommand{\bR}{\mathbb{R}}

\newcommand{\cD}{\mathcal{D}}

\newcommand{\tot}{\mathsf{tot}}
\newcommand{\feat}{\mathsf{feat}}

\newcommand{\dset}{\mathsf{set}}
\newcommand{\train}{\mathsf{train}}
\newcommand{\validate}{\mathsf{validate}}
\newcommand{\test}{\mathsf{test}}

\newcommand{\metric}{\mathsf{metric}}
\newcommand{\bh}{\mathsf{bh}}
\newcommand{\stockout}{\mathsf{SR}}
\newcommand{\turnovertime}{\mathsf{TT}}

\newcommand{\DRL}{\textsc{DRL}\xspace}
\newcommand{\DDPG}{\textsc{DDPG}\xspace}
\newcommand{\PPO}{\textsc{PPO}\xspace}
\newcommand{\DiffSimu}{\textsc{DS}\xspace}

\newcommand{\Reg}{\textsc{Reg}\xspace}
\newcommand{\None}{\textsc{None}\xspace}
\newcommand{\Base}{\textsc{Base}\xspace}
\newcommand{\Coeff}{\textsc{Coeff}\xspace}
\newcommand{\Both}{\textsc{Both}\xspace}

\usepackage{color}              %
\usepackage{color-edits}

\usepackage[english]{babel}
\usepackage[autostyle, english = american]{csquotes}
\MakeOuterQuote{"}

\usepackage{natbib}
 \bibpunct[, ]{(}{)}{,}{a}{}{,}%
 \def\bibfont{\small}%

\EquationsNumberedThrough    %

\TheoremsNumberedThrough     %
\ECRepeatTheorems  %

\begin{document}

\RUNTITLE{DeepStock: Reinforcement Learning with Policy Regularizations for Inventory Management}
\TITLE{DeepStock: Reinforcement Learning with Policy Regularizations for Inventory Management}

\ARTICLEAUTHORS{%
\AUTHOR{
\bf Yaqi Xie\textsuperscript{1},
Xinru Hao\textsuperscript{2},
Jiaxi Liu\textsuperscript{3},
Will Ma\textsuperscript{4},
Linwei Xin\textsuperscript{5},
Lei Cao\textsuperscript{2},
Yidong Zhang\textsuperscript{2}
}
\AFF{\textsuperscript{1}Booth School of Business, University of Chicago, Chicago, USA. \texttt{yaqi.xie@chicagobooth.edu}}%
\AFF{\textsuperscript{2}Taobao \& Tmall Group, Hangzhou, China. \texttt{xinru.hxr@taobao.com;huaju.cl@taobao.com;yidongzster@gmail.com}}%
\AFF{\textsuperscript{3}School of Economics, Sichuan University, Chengdu, China. \texttt{liujiaxi@stu.scu.edu.cn}}%
\AFF{\textsuperscript{4}Graduate School of Business, Columbia University, New York, USA. \texttt{wm2428@gsb.columbia.edu}}%
\AFF{\textsuperscript{5}School of Operations Research and Information Engineering, Cornell University, Ithaca, USA. \texttt{lx267@cornell.edu}}%

\RUNAUTHOR{Xie et al.}
}

\ABSTRACT{Deep Reinforcement Learning (DRL) provides a general-purpose methodology for training inventory policies that can leverage big data and compute.
However, off-the-shelf implementations of DRL have seen mixed success, often plagued by high sensitivity to the hyperparameters used during training. In this paper, we show that by imposing policy regularizations, grounded in classical inventory concepts such as "Base Stock", we can significantly accelerate hyperparameter tuning and improve the final performance of several DRL methods. We report details from a {100\%} deployment of DRL with policy regularizations on Alibaba's e-commerce platform, Tmall. We also include extensive synthetic experiments, which show that policy regularizations reshape the narrative on what is the best DRL method for inventory management.

}

\HISTORY{Final version; forthcoming in \textit{INFORMS Journal on Applied Analytics}}
\maketitle

\footnotetext{YX, XH, and JL are co-first authors with equal contribution. WM and LX provided academic guidance, while LC and YZ offered industry leadership.}

\vspace{-1cm}

\section{Introduction}\label{sec:intro}

Inventory control is a foundational problem in the field of Operations Research. Historically, the focus has been on stylized inventory and demand models that allow for the derivation of optimal policies.
In modern times, these models can be significantly enhanced by leveraging high-dimensional contextual data. Even though optimal policies are no longer analytically defined, Deep Reinforcement Learning (DRL) offers a promising general-purpose methodology for learning performant policies that act on this high-dimensional data.

However, the success of applying DRL to inventory control has been mixed \citep[see][]{gijsbrechts2025ai}. Indeed, a common approach is to frame inventory control as a generic sequential decision-making problem and then apply off-the-shelf DRL methods, which leads to actions that are difficult to interpret. More importantly, a dealbreaker is that the performance of DRL is highly sensitive to hyperparameters, whose tuning is time-consuming.

In this paper, we show how these dealbreakers can be mitigated through policy regularizations, where we encode simple intuitions from inventory theory into the policy learned by DRL. One such regularization incorporates the inventory concept of "Base Stock" into Deep RL, which is why we call our algorithm "DeepStock". This not only reduces the black-box nature of DRL, but drastically speeds up training and improves final performance.
In fact, our policy regularization techniques have enabled the full-scale deployment of DRL at Alibaba. As of October 2025, our algorithm manages inventory replenishment for 100\% of the products (both domestic and international) sold by Alibaba on its B2C e-commerce platform, Tmall, covering over 1 million SKU-warehouse combinations.

\subsection{Description of DRL and Policy Regularizations}

A DRL method learns a policy $\pi$, represented by a (deep) neural network, that can decide an inventory ordering action from any state.
In our problem, the state of an SKU at a time $t$ includes exogenous features $x_t\in\bR^m$, which consists of both
static attributes (e.g., product category, demand magnitude, supplier, lead time, review period, profit margin) and
dynamic attributes that evolve over time (e.g., upcoming promotions, seasonality, recent social media trends).
In addition, the state includes endogenous information $I_t$ about upcoming inventory shipments, which depend on prior actions (i.e., previously placed orders).
This rich state space allows for meta-learning across SKU's with diverse characteristics, such as in demand magnitude (e.g., 10 vs.\ 10,000 weekly sales), lead time (domestic vs.\ international), and general demand pattern (e.g., fast-moving vs.\ long-tail).
Neural networks can encode this high-dimensional information into meaningful representations while filtering out noisy signals.

When a DRL algorithm is applied off-the-shelf, the policy $\pi$ is a typically single neural network that outputs an order quantity $\pi(I_t,x_t)$ based on the input state $I_t,x_t$.
In the following, our policy regularizations define a structured mapping from the neural network output to an order quantity, with each regularization providing a specific interpretation of the network output.
Importantly, because the neural network can in principle learn to undo these mappings, the regularizations do not restrict the expressiveness of the policy class. They are better understood as an \textit{action remapping} that provides a more natural parameterization for inventory decisions, biasing the learning process toward sensible behavior without constraining the space of achievable policies.

Our first policy regularization imposes that the order quantity takes the functional form $\pi(I_t,x_t)=\max \left\{\mu_\Base(I_t,x_t)-\tot(I_t),0\right\}$, where $\mu_\Base$ is the learned neural network, and $\tot$ denotes the total inventory (including upcoming shipments) contained in $I_t$.
Here, $\Base$ stands for "base stock", where $\mu_\Base(I_t,x_t)$ represents the target level for the total inventory, and the order quantity $\pi(I_t,x_t)$ equals this target level minus the total inventory $\tot(I_t)$ we already have.  Intuitively, the learned target $\mu_\Base(I_t,x_t)$ should depend mostly on the exogenous features $x_t$ which forecasts the upcoming demand, while the inventory information $I_t$ is incorporated into the decision through the $\tot(I_t)$ term.

Our second policy regularization imposes the functional form $\pi(I_t,x_t)=\mu_\Coeff(I_t,x_t)^\top \feat(x_t)$, where $\mu_\Coeff$ is a learned neural network with an $m'$-dimensional output, providing Coefficients for $m'$ features extracted from $x_t\in\bR^m$ by the mapping $\feat(x_t)\in\bR^{m'}$.
At Alibaba, we have $m'=5$, with $\feat(x_t)$ consisting of short- and long-term historical demands, short- and long-term forecast demands, and a constant bias term.
These features capture distinct demand patterns and signals across historical versus forecasted information and near versus distant horizons.
Intuitively, the desired order quantity should exhibit a positive relationship with these features.
Moreover, multiplying the learned coefficients by these demand-related features helps rescale, or denormalize, the resulting action with respect to the SKU’s demand magnitude, thereby stabilizing meta-learning across SKU's with different demand magnitudes.

Finally, we can combine both of our policy regularizations, in which case we impose the order quantity to take the functional form $\pi(I_t,x_t)= \max \left\{ \mu_\Both(I_t,x_t)^\top \feat(x_t)-\tot(I_t), 0\right\}$, with $\mu_\Both$ being the learned neural network with an $m'$-dimensional output.

\subsection{Main Results}

Our policy regularizations can be tested in conjunction with any DRL method for training the neural network that defines the policy.
We test two traditional DRL methods,
{Deep Deterministic Policy Gradient (\DDPG) \citep{lillicrap2015continuous}  and Proximal Policy Optimization (\PPO) \citep{schulman2017proximal}}, which are representative model-free DRL methods that span different facets of DRL.

\DDPG and \PPO may only perform well under specific hyperparameter configurations, which change with each dataset, causing a major bottleneck to their deployment in the real world.  In fact, other e-commerce firms like Amazon have adopted a different DRL method that directly takes gradients of the total simulated loss over a specified time horizon \citep{madeka2022deep}, instead of learning Cost-to-go functions for intermediate states which is key to RL.
The performance of this "Differentiable Simulator" (\DiffSimu) method is much less sensitive to hyperparameter tuning.

We test our policy regularizations with \DDPG, \PPO, and \DiffSimu, and report results based on synthetic data, offline data from Alibaba, and real-world deployments at Alibaba.

\begin{center}
\begin{tcolorbox}[colback=gray!20, colframe=white, width=\textwidth, boxrule=0pt, arc=2mm, auto outer arc,    after=\vspace{-1pt} ]
\textbf{Takeaway~I}: Policy regularizations improve the performance of DRL methods, \textit{especially when hyperparameter tuning is limited}.
\end{tcolorbox}
\end{center}

We demonstrate Takeaway~I on Alibaba's offline data and especially our synthetic data, where we plot the best-so-far performance of DRL methods, with and without regularization, after each hyperparameter trial. The improvement from regularization is more drastic under a limited number of hyperparameter trials, even though the best performance after many trials also improves for all of \DDPG, \PPO, and \DiffSimu.

To explain this finding, our regularizations encode into the policy human intuitions that prevent obvious blunders: our \Base regularization discourages large orders when we already have a lot of inventory in $I_t$ and ensures a stable total inventory, while our \Coeff regularization ensures that larger historical and forecasted demands imply larger orders.
This especially improves the policy learned under the untuned hyperparameter configurations,
where learning may be hindered by obvious blunders.
Although the policy regularizations do not technically restrict the policy space, our synthetic experiments demonstrate that they significantly accelerate Q-function convergence in DRL.

\begin{center}
\begin{tcolorbox}[colback=gray!20, colframe=white, width=\textwidth, boxrule=0pt, arc=2mm, auto outer arc,     after=\vspace{-1pt}]
\textbf{Takeaway~II}: Policy regularizations \textit{tilt the scales} in favor of traditional DRL methods (\DDPG, \PPO) over \DiffSimu.
\end{tcolorbox}
\end{center}

We demonstrate Takeaway~II on Alibaba's offline data and our synthetic data.
As explained earlier, policy regularizations improve both traditional DRL and \DiffSimu, but the improvement is more significant when hyperparameter tuning is limited and when learning a Q-function, which is more likely to be the case for traditional DRL than for \DiffSimu.

We also dig deeper in our synthetic experiments to explain why traditional DRL can potentially beat \DiffSimu.
In our synthetic settings, we find that \DiffSimu has a tendency to overfit, because it does not explicitly learn intermediate structure in the form of a Cost-to-go function from a state or a Q-function from a state-action pair.
However, given enough IID trajectories describing the evolution of SKU features and demands, the overfitting effect diminishes, and \DiffSimu performs no worse than traditional DRL (with or without regularization).
Separately, Alibaba also found \DDPG to outperform \DiffSimu when learning from 55,000 90-day SKU trajectories on its offline data, although the reasons may differ given the much higher dimensionality of the real-world setting.

\begin{center}
\begin{tcolorbox}[colback=gray!20, colframe=white, width=\textwidth, boxrule=0pt, arc=2mm, auto outer arc, after=\vspace{-1pt}]
\textbf{Takeaway~III}: Policy regularizations enable a \textit{unified, full-scale} deployment of DRL.
\end{tcolorbox}
\end{center}

We demonstrate Takeaway~III by reporting the impact of policy regularizations on Alibaba's deployment of \DRL in the real world.
Alibaba proceeded to deploy \DDPG with our policy regularizations, which generally achieved the best performance in its routine offline tests, as exemplified by the results on Alibaba's offline data presented in this paper.
Although Alibaba has previously tinkered with DRL as reported in \citet{liu2023ai}, that work primarily addressed uncertain order yields arising from supply shortages during the COVID-19 pandemic. In addition, that DRL version was not general-purpose: it required grouping similar products and training separate models per group.
By contrast, our policy regularizations have allowed for a full-scale deployment of DRL that is \textit{unified}, where we can meta-learn a single policy across all SKU's by leveraging their features, instead of separating them into groups such as fast-moving vs.\ long-tail SKU's. %

It is worth emphasizing that both the dynamic attributes in $x_t$ and the inventory state $I_t$ are normalized across time for each SKU before being fed into the neural network. In conjunction with our \Coeff regularization, this facilitates meta-learning across SKU's.  To explain why, the normalization allows us to interpret patterns in dynamic attributes and inventory states in a consistent manner across SKU's with different sales magnitudes, while the multiplication by $\feat(x_t)$ in our \Coeff regularization ensures that the resulting actions are properly denormalized with respect to the demand magnitudes.
This allows unified meta-learning across all SKU's, without requiring explicit clustering.

\paragraph{Paper outline.} This rest of paper contains the following sections:
State of the Literature on RL for Inventory;
Inventory Model, Metrics, and Policies;
Training using DRL;
Experiments on Synthetic Data;
Experiments on Alibaba's Offline Data;
Real-world Deployment and Impact at Alibaba;
Conclusion.

\section{State of the Literature on RL for Inventory}

\paragraph{Policy regularizations.}
To the best of our knowledge, simple forms of policy regularizations such as the "base stock" structure have not been previously explored in the RL literature for inventory management. Our high-level approach is to minimize an empirical loss function while imposing a policy structure that is satisfied by the optimal policy under stylized models, which does align with the sample complexity theory developed in \citet{xie2024vc}.

Many previous works however have encoded inventory policy structures into RL. As examples, \cite{de2022reward} modify the reward function by embedding structure from heuristic inventory policies, using those as teacher policies; \cite{qi2023practical} use a labeling method that captures the behavior of the optimal dynamic programming solution under historical observations, and then use that as a regularizer on the learned policy; \cite{maggiar2025structure} impose a penalty term in the objective when the learned policy violates certain structural properties known to hold for optimal policies (as proved in the literature). These works generally differ from ours by incorporating penalty terms into the objective function, rather than directly restricting the policy structure.
A related work that adopts the base-stock policy class is \cite{vanvuchelen2025use}. However, they study a non-contextual joint replenishment setting, where the state consists of the inventory levels of all SKU's to capture interdependencies from shared ordering costs. In contrast, we consider a high-dimensional contextual setting with a large number of SKU's, where each SKU’s decision depends on its own contextual features.

Finally, we note that
our policy regularizations are problem-specific
and should not be confused with generic regularization techniques commonly used in RL, such as entropy regularization \citep{haarnoja2018soft} or those used in trust-region policy optimization \citep{schulman2015trust}.

\paragraph{DS vs.\ traditional DRL for inventory.}
Although \citet{madeka2022deep} and concurrently \citet{alvo2023neural} have been promoting \DiffSimu when doing DRL for inventory management, our paper aims to compare \DiffSimu to traditional DRL with proper hyperparameter tuning, finding that traditional DRL (with policy regularizations) can actually outperform.
Our synthetic experiments suggest a potential shortcoming of \DiffSimu: its inability to learn across time when there are not enough parallel trajectories.
That being said, \DiffSimu performs excellently in synthetic settings given enough IID trajectories, which is consistent with the experimental findings in \citet{alvo2023neural}.

\paragraph{Large-scale deployments of DRL for inventory.} There has been increasing documentation of deep learning deployments in inventory management at leading e-retailers, including Alibaba \citep{liu2023ai}, Amazon \citep{madeka2022deep}, and JD.com \citep{qi2023practical}. A distinction of our work is that we achieve 100\% coverage of all products sold by Alibaba on its Tmall e-commerce platform, which is only possible by training a unified DRL policy. Although all these works (including ours) omit many implementation details due to corporate confidentiality,
our work also includes synthetic experiments that aim to academically illustrate our main findings inside the company.

\section{Inventory Model, Metrics, and Policies}

\paragraph{Inventory dynamics.}
Let $T$, $P$, and $L$ be positive integers.
 We consider a discrete-time horizon of $t = 1, \ldots, T$ days.
 Inventory orders are placed every $P$ days starting at the beginning of day $P+1$ (the ``Period''), specifically at the beginnings of days $P+1, 2P+1, 3P+1, \ldots$, with the first $P$ days serving as a warm-up period.
 Ordered inventory arrives $L$ days later (the ``Lead time''), i.e., at the beginnings of days $P+1+L, 2P+1+L, 3P+1+L, \ldots$
We let $I_t=(I^0_t,I^1_t,\ldots,I^{L-1}_t)$ denote the inventory state vector at the beginning of day $t$, where $I^0_t$ denotes the inventory on hand, and $I^\ell_t$ denotes the inventory that will arrive $\ell$ days into the future, for all $\ell=1,\ldots,L-1$.
We initialize $I^0_1=I^1_1=\cdots=I^{L-1}_1=0$.

If inventory is to be ordered on day $t$, we decide the quantity $I^L_t$ at the beginning of day $t$; otherwise we set $I^L_t=0$.
Over the course of the day, $d_t$ consumers each attempt to purchase one unit of inventory, where $d_t$ is called the "demand". The actual sales made are $\min\{I^0_t,d_t\}$, and the inventory vector for the next day is updated as
\begingroup
\setlength{\abovedisplayskip}{3pt}
\setlength{\belowdisplayskip}{2pt}
\begin{align} \label{eqn:invUpdate}
I_{t+1}^0 = I^0_t-\min\{I^0_t,d_t\}+I^1_t= \max\{I_t^0-d_t, 0\}+I^1_t;
\qquad I^1_{t+1}=I^2_t,\ldots, I^{L-1}_{t+1}=I^L_t.
\end{align}
\endgroup

\paragraph{Performance metrics.}
In inventory theory, the standard loss objective to minimize is the underage and overage loss
\begingroup
\setlength{\abovedisplayskip}{2.5pt}
\setlength{\belowdisplayskip}{2.5pt}
\begin{align*}%
\ell_\bh:=\sum_{t=P+L+1}^T \left(b\cdot\max\{d_t-I^0_t,0\} + h\cdot\max\{I^0_t-d_t,0\}\right).
\end{align*}
\endgroup
That is, at the end of each day $t$ (excluding the first $P+L$ days before the initial order arrives): if $d_t\ge I^0_t$ (i.e., the inventory stocked out), then we are penalized $b$ times the unmet demand $d_t-I^0_t$; otherwise, we are penalized $h$ times the leftover inventory $I^0_t-d_t$.

However, $\ell_\bh$ is not a practical evaluation metric, as quantifying $b$ and $h$ is difficult in practice, particularly since unmet demand during stockouts cannot be directly observed when $d_t$ is unknown. Therefore, at Alibaba, we evaluate instead the following two metrics:
\begin{align*}%
\ell_\stockout := \frac{1}{T-P-L}\sum_{t=P+L+1}^T \bI (d_t\ge I^0_t )\quad\text{and}\quad
\ell_\turnovertime := \frac{\frac{1}{T-P-L} \sum_{t=P+L+1}^T \max\{I^0_t-d_t,0\}}{ \frac{1}{T-P-L}
 \sum_{t=P+L+1}^T \min\{I_t^0, d_t\} }.
\end{align*}
Here, the "Stockout Rate" loss $\ell_\stockout$ measures the \% of days ending in stockout, penalizing having too little inventory.
On the other hand, the "Turnover Time" loss $\ell_\turnovertime$ computes the average end-of-day inventory divided by average sales, which measures the average duration that a unit of inventory stays on hand, penalizing having too much inventory.

In our synthetic experiments, we train and evaluate policies using the standard objective $\ell_\bh$. We note that the loss objective $\ell_\bh$ is naturally additive over time $t$, providing a reward signal for each action taken during RL training.
At Alibaba, the policies are trained and evaluated using a weighted combination of $\ell_\stockout$ and $\ell_\turnovertime$.
We can adjust the relative weight to trade off between $\ell_\stockout$ and $\ell_\turnovertime$ in practice, ideally getting a Pareto improvement where both $\ell_\stockout,\ell_\turnovertime$ decrease.
Although $\ell_\turnovertime$ cannot be exactly decomposed into per-period losses, Alibaba approximates the immediate loss of an action by applying the Turnover Time loss over the periods from its arrival until the arrival of the next action.

\paragraph{Contexts and policies.}
For all $t=1,\ldots,T$, we let $x_t\in\bR^m$ denote a context vector available at the beginning of day $t$, that may be used to inform the inventory ordering decision.
We assume that $x_t$ contains all the up-to-date information and repeats static information such as $P$ and $L$.
An inventory policy $\pi$ outputs an order quantity given any inventory vector $I_t$ and context vector $x_t$, setting the decision variable $I^L_t$ to $\pi(I_t,x_t)$ (i.e., $\pi$ does not depend on $x_{t-1}$, as that information is already outdated).
The order quantity $\pi(I_t,x_t)$ is assumed to lie in a bounded continuous space.
We assume without loss that $\pi$ is stationary, i.e.,\ does not depend on $t$, because any time-specific information (e.g., holidays or seasonality) can be encoded into the context vector $x_t$.

\paragraph{Discussion of model assumptions.}
We will train inventory policies $\pi$ by simulating the above inventory dynamics using historical trajectories  $\xi=(x_t,d_t)_{t=1}^T$.
We emphasize that the trained policies will be able to output decisions in real-world environments even if these simplified model dynamics are violated.
For example, we relax order quantities to a continuous space during training, although real-world inventory decisions are typically integer-valued. Similarly,
if we want to make an emergency order on a day $t\neq P+1, 2P+1, 3P+1\ldots$, then we can still evaluate $\pi(I_t,x_t)$ based on the current $I_t$ and $x_t$ to get an ordering decision; meanwhile, if there are unexpected shipping delays, then the inventory update rule would deviate from~\eqref{eqn:invUpdate} but the policy would proceed all the same.
Of course, the resulting decisions may perform poorly under model violations---for example, if the actual lead time is highly random, then a policy calibrated for a deterministic $L$ may fail to maintain sufficient inventory on hand. That being said, our success in real-world environments suggests that the policy trained on this simplified model is sufficient, and our general experience is that being able to train better because the model is simpler outweighs the benefits of modeling more complex inventory dynamics.

\section{Training using Deep Reinforcement Learning (DRL)}

We generally let $\cD$ denote a dataset of trajectories $\xi=(x_t,d_t)_{t=1}^T$,
which can be synthetically generated, or based on historical data internal to Alibaba.
In Alibaba's data, $d_t$ must be inferred if there was a stockout during day $t$, and this can be done by extrapolating $d_t$ based on time of stockout.
We train using these inferred demands, and again, the success of the trained policy after deployment suggests that the extrapolation procedure was appropriate.

We distinguish between datasets used for training, validation, and testing, denoting them using $\cD^\train$, $\cD^\validate$, and $\cD^\test$, respectively.
$\cD^\train\cup\cD^\validate$ is the data available for learning $\pi$, where during training $\pi$ is updated iteratively based on the trajectories in $\cD^\train$, and a separate dataset $\cD^\validate$ is used to evaluate intermediate policies learned during the training and determine when training should stop.
Finally, a third dataset $\cD^\test$ is used to evaluate out-of-sample performance, although in deployment we would evaluate based on the actually observed $\ell_\stockout$ and $\ell_\turnovertime$ metrics.

For any $\dset\in\{\train,\validate,\test\}$ and $\metric\in\{\bh,\stockout,\turnovertime\}$, we define
\begingroup
\setlength{\abovedisplayskip}{2.5pt}
\setlength{\belowdisplayskip}{2pt}
\begin{align*}
L^\dset_\metric(\pi):=\frac{1}{|\cD^\dset|}\sum_{\xi\in\cD^\dset}\ell_\metric(\pi,\xi),
\end{align*}
\endgroup
which is the average of loss $\ell_\metric(\pi,\xi)$ over the trajectories $\xi\in\cD^\dset$, for a policy $\pi$.  %

\subsection{Overview of DRL Training Methods}

A DRL method learns a policy $\pi$ based on $\cD^\train\cup\cD^\validate$.
While there is an enormous variety of DRL methods \citep[see e.g.,][]{achiam2018}, we focus on \textit{model-free} methods which directly minimize the loss of $\pi$ on trajectories in $\cD^\train\cup\cD^\validate$, instead of trying to learn a stochastic model for the evolution of $\xi=(x_t,d_t)_{t=1}^T$ and then using that to optimize $\pi$.
Off-policy model-free DRL can learn purely by observing historical policies; meanwhile, on-policy model-free DRL requires simulating the current policies, which we can do, by replaying historical trajectories $\xi=(x_t,d_t)_{t=1}^T$ with the inferred demands $d_t$ on our inventory model.
Key concepts in model-free DRL include taking policy gradients and learning Q-functions; the methods we implement consider both of these concepts.

In this paper we consider
{\DDPG and \PPO}, which we believe to be appropriate for our inventory problem, and also span these distinctions between DRL methods.
In particular, \PPO simulates a randomized $\pi$ and reinforces actions that generate high rewards under on-policy trajectories, being a modern policy gradient method that "clips" updates to prevent $\pi$ from changing too quickly. \PPO can include additional tricks such as generalized advantage estimation \citep{schulman2015high} and an entropy bonus to encourage exploration.
Meanwhile, \DDPG learns a Q-function from a buffer of off-policy observations, which is used to optimize a (deterministic) policy by taking gradients of the Q-function in the continuous action space (as is the case in our inventory problem).
Given that on-policy simulation is also possible in our inventory problem, we always add the observations from the most recent policy to the buffer for learning the Q-function.
\DDPG can include additional tricks such as using a target network to stabilize the policy and Q-function updates.

Finally, we consider \DiffSimu, the differentiable simulator method which takes policy gradients based on loss on the entire trajectory.  It fully exploits the fact that any counterfactual inventory policy can be simulated on historical trajectories while allowing gradients.

\paragraph{Hyperparameter tuning.}
For any DRL method, the specific update rule for $\pi$ depends on hyperparameters (e.g., "learning rate", which determines the step size of each update).
A DRL method may perform well only under carefully tuned hyperparameter configurations, which can vary across different datasets.
Finding the optimal configuration is challenging, as both \DDPG and \PPO have at least 10 hyperparameters that may need to be adjusted to achieve good performance, and a configuration can only be evaluated after the training run converges, which can take many back-and-forths between training and validation.
Exacerbating the challenge, even for a fixed hyperparameter configuration the performance of a training run is random, depending on the seed.
We follow a standard hyperparameter tuning procedure that first uses Bayesian optimization to search over configurations, and then runs the most promising configurations with multiple seeds each to determine a winner.

Further details about our implementations of \DDPG, \PPO, and \DiffSimu, including hyperparameter search ranges, can be found in Appendix~A.

\subsection{Combining DRL Methods with Policy Regularizations}

A DRL method specifies how to train $\pi$, which can be used in conjunction with any of our policy regularizations (detailed in the Introduction) that impose a functional form on $\pi$.
For each combination of $\DRL\in\{\DDPG,\PPO,\DiffSimu\}$ and $\Reg\in\{\None,\Base,\Coeff,\Both\}$ (where \None refers to "no regularization", i.e.,~$\pi$ directly outputs an order quantity), we let $\pi^{\DRL,\Reg}$ denote the final policy learned.

\section{Experiments on Synthetic Data}
In this section, we present synthetic experiments to validate our results. The code is publicly available at \url{https://github.com/xieyaqi188/DRL_inventory_Alibaba}.
We fix $P=2,L=1,m=1$, and set $T=31$ unless stated otherwise. While the lost-sales dynamics introduced in \eqref{eqn:invUpdate} are common in practice (e.g., at Alibaba), we consider the backlogged dynamics $I_{t+1}^0 = I^0_t- d_t +I^1_t$ (i.e., $I_{t+1}^0$ can be negative) in the synthetic experiments, as the optimal policy is relatively easy to compute. This formulation is also standard in many stylized inventory models and complements the lost-sales dynamics.

For each setting, we construct datasets $\cD^\train,\cD^\validate,\cD^\test$ by drawing $\xi=(x_t,d_t)_{t=1}^T$ from trajectory distributions. %
In our first set of trajectory distributions which we call INDEP, for all $t=1,\ldots,T$, the feature $x_t$ is set to be $\left\lfloor \frac{t-1}{2}\right\rfloor$, and $d_t$ is drawn from a scalar distribution that depends on $t$ but is independent across $t$. %
We also consider trajectory-generation schemes in which the demand distribution for each SKU in $\cD^\train$, $\cD^\validate$, or $\cD^\test$ is determined by parameters initialized randomly drawn from a prior, capturing the varying demand magnitudes observed across SKU's. Indeed, in our second set of trajectory distributions, $d_t$ is correlated over time following an AR(1) stochastic process with the drawn parameters, and the feature is $x_t=d_{t-1}$ which determines the state of the stochastic process (if the parameters were known).
In our third set of trajectory distributions, $d_t$ is IID across $t$ following a distribution determined by a single parameter, and the feature is set exactly to this parameter.
Details of these trajectory distributions can be found in Appendix~A.
We construct datasets for the following settings:
\begin{enumerate}
\item INDEP distribution with 20 sample trajectories, split so that $|\cD^\train|=|\cD^\validate|=10$;
\item AR(1) distribution with 20 sample trajectories, split so that $|\cD^\train|=|\cD^\validate|=10$;
\item AR(1) distribution with 10 sample trajectories, split so that $|\cD^\train|=|\cD^\validate|=5$;
\item IID distribution with sample sizes $|\cD^\train|=|\cD^\validate|\in \{5, 10, 20\}$ and horizon lengths $T\in \{7, 19, 35, 67, 131\}$.
\end{enumerate}
For the first three settings, we fix $|\cD^\test|=100$ for evaluation. In the fourth setting, we assume there are $|\cD^\train|$ SKU's (with distinct parameters) in total, and we set $\cD^\train$, $\cD^\validate$ and $\cD^\test$ to include 1, 1, and 100 trajectories per SKU, respectively (i.e., $|\cD^\test|=100 \times |\cD^\train|$).

We compare policies $\pi^{\DRL,\Reg}$ for combinations of $\DRL\in\{\DDPG,\PPO,\DiffSimu\}$ and $\Reg\in\{\None,\Base\}$, omitting our \Coeff regularizations because $x_t$ is one-dimensional.
We train, validate, and test using the same $\ell_{\bh}$ metric, with $b=0.9$ and $h=0.1$.
During training of \DDPG and \PPO, the immediate loss of an action is  computed by applying $\ell_{\bh}$ over the periods from its arrival until the arrival of the next action.

\begin{figure}
\begin{center}
\footnotesize{Setting 1 (INDEP trajectory distribution, 20 samples)}

\includegraphics[width=0.75\linewidth]{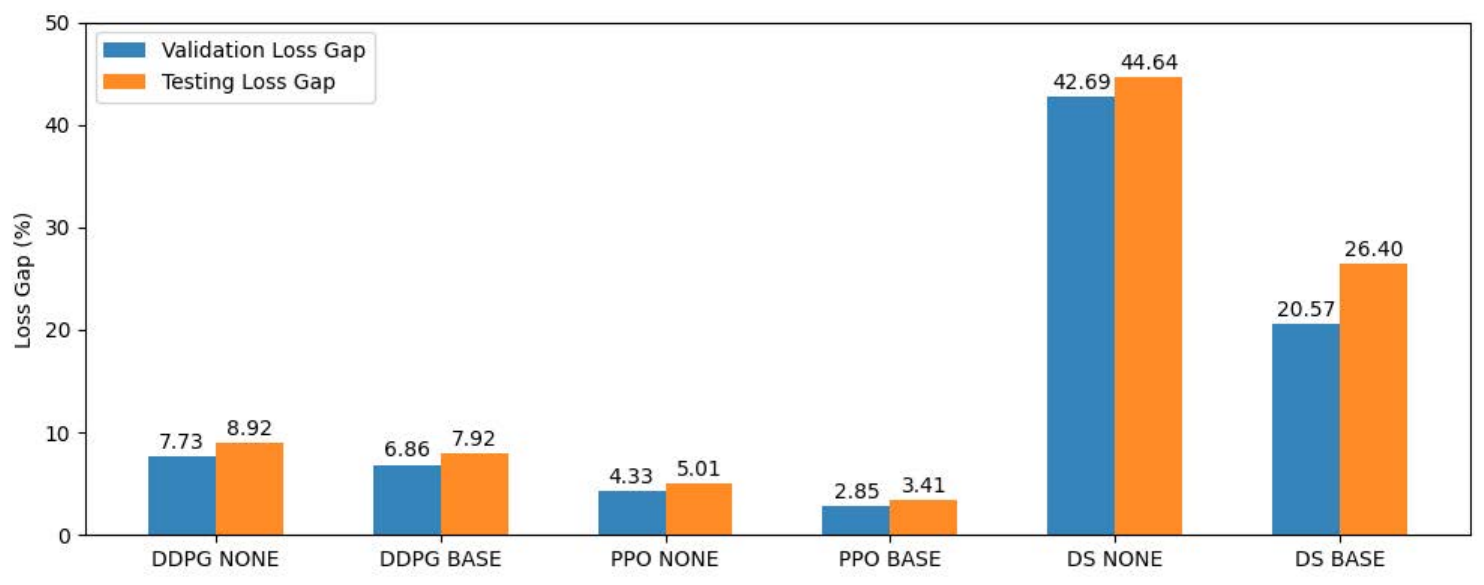}
\end{center}

\begin{center}
\footnotesize{Setting 2 (AR(1) trajectory distribution, 20 samples)}

\includegraphics[width=0.75\linewidth]{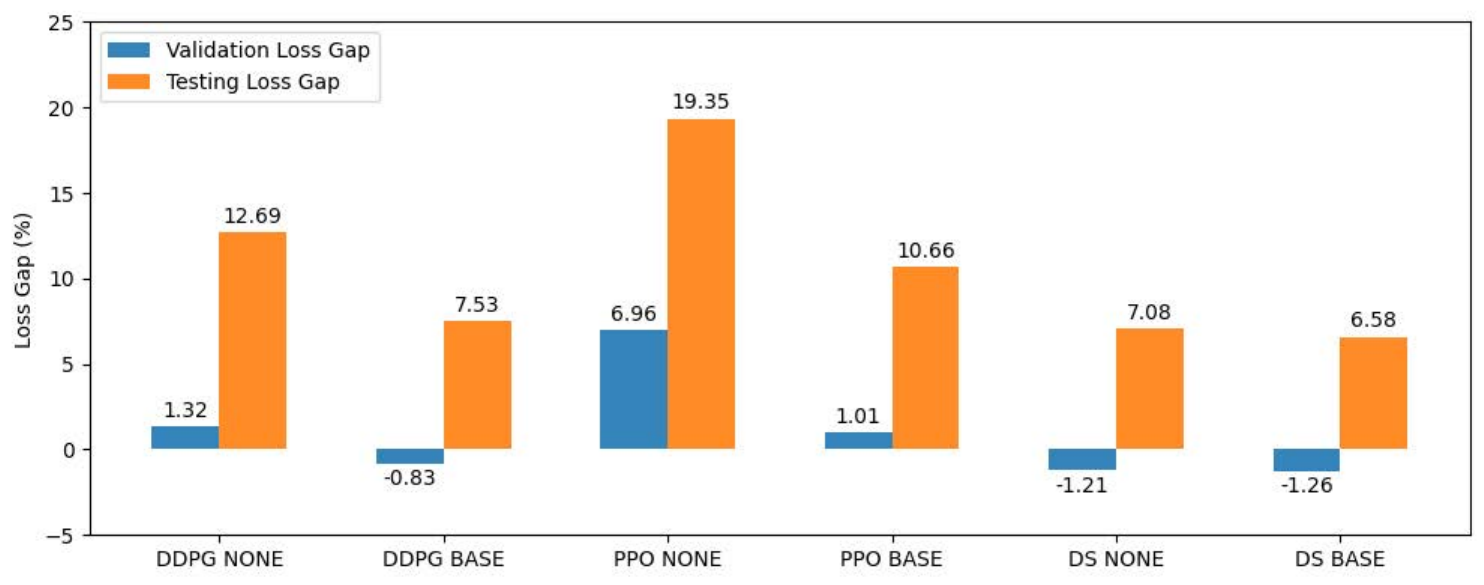}
\end{center}

\begin{center}
\footnotesize{Setting 3 (AR(1) trajectory distribution, 10 samples)}

\includegraphics[width=0.75\linewidth]{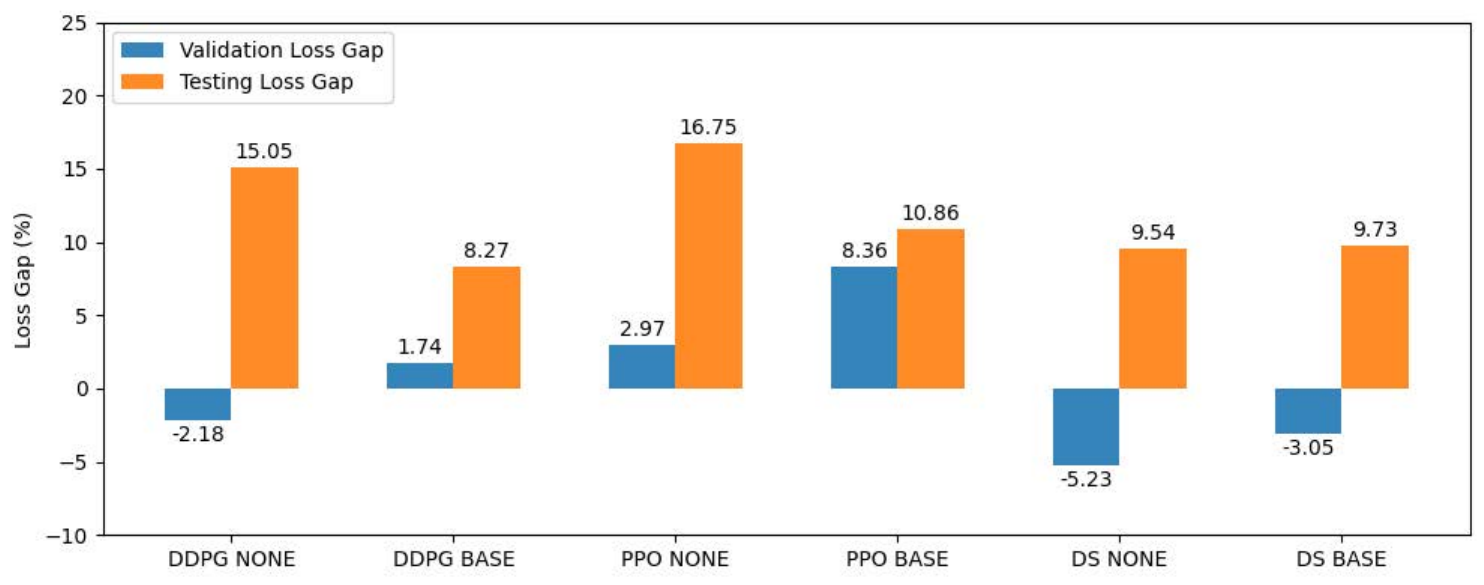}
\end{center}
\caption{Validation and Testing Loss Gaps for the 6 combinations of DRL Method and Policy Regularization.}
\label{fig-exp1-valtest}
\end{figure}

We consider the performance of $\pi^{\DRL,\Reg}$ relative to $\pi^*$, which we use to denote the optimal policy knowing the true trajectory distribution.  For our AR(1) distribution, we approximate $\pi^*$ by optimizing over the $\cD^\test$ dataset.
In either case, we report
$$
\text{Validation Loss Gap}:=\frac{L^\validate_\bh(\pi^{\DRL,\Reg})}{L^\test_\bh(\pi^*)}-1
\quad\text{and}\quad
\text{Testing Loss Gap}:=\frac{L^\test_\bh(\pi^{\DRL,\Reg})}{L^\test_\bh(\pi^*)}-1.
$$
The Testing Loss Gap is our main metric of interest, which is out-of-sample and always non-negative, although we also include the Validation Loss Gap, which is in-sample and can be negative.
For each combination of $\DRL$ and $\Reg$, we report averages from running the hyperparameter tuning procedure three times, for each of {Settings~1--4}, in \Cref{fig-exp1-valtest}.

\begin{figure}
\centering
\includegraphics[width=0.8\linewidth]{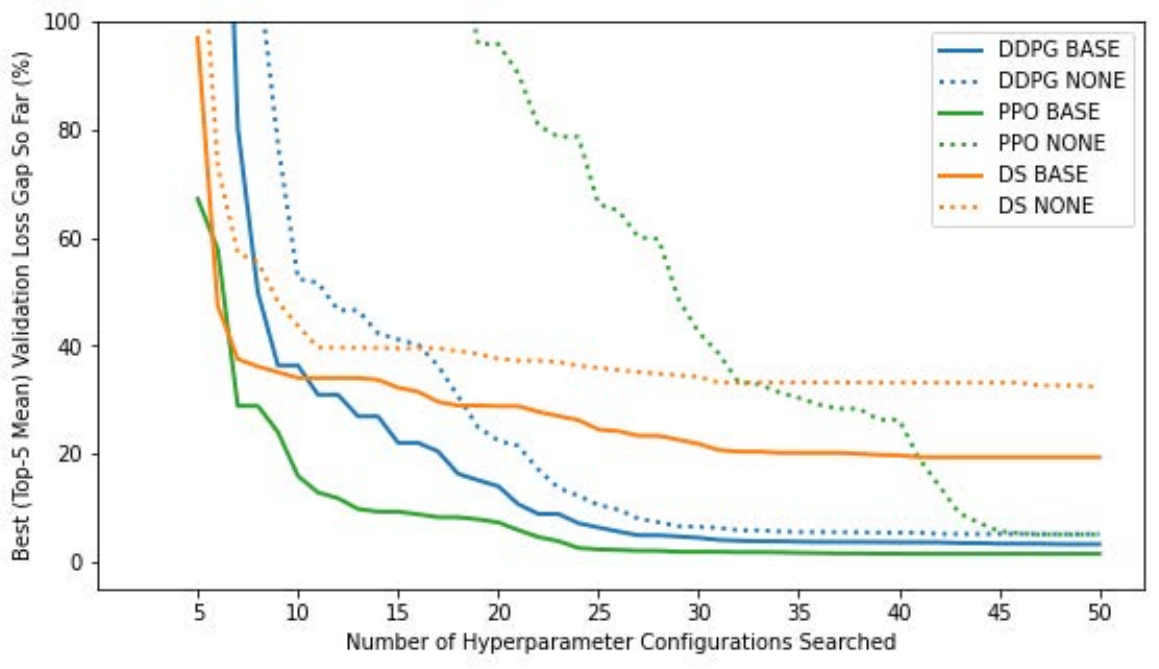}
\caption{Validation Loss Gaps for the top-5 hyperparameter configurations, shown for each combination of DRL Method and Policy Regularization, in Setting 1.
}
\label{fig-exp1-hyperpara}
\end{figure}

\paragraph{Results.}
As seen in \Cref{fig-exp1-valtest},
the \Base regularization improves the Testing Loss of the \DRL\ method
in 8 out of 9 cases (all except for \DiffSimu in Setting~3), often significantly.
Furthermore, \Cref{fig-exp1-hyperpara} shows the best-so-far performance\footnote{Technically this is Validation not Testing Loss, because we do not compute the latter during hyperparameter tuning.  However, \Cref{fig-exp1-valtest} showed that the distinction between Validation and Testing Loss is unimportant in Setting~1.} in Setting~1 as hyperparameter trials are being conducted, for each $\pi^{\DRL,\Reg}$. We can see for all three \DRL methods that the improvement from $\Reg=\None$ to $\Reg=\Base$ is greater under limited hyperparameter search, corroborating our Takeaway~I from the Introduction.

Still on \Cref{fig-exp1-hyperpara}, we observe that without regularization, \DDPG needs to search 17 configurations to surpass \DiffSimu, while \PPO needs to search 32. By contrast, with the \Base regularization, these numbers reduce to 11 and 7, respectively.
This suggests that policy regularization greatly reduces the hyperparameter tuning effort required for traditional DRL methods (\DDPG, \PPO) to outperform \DiffSimu.
Returning to \Cref{fig-exp1-valtest}, we observe that regularization also improves the final Testing Loss (after hyperparameter search is finished) of \DDPG and \PPO by a greater amount than \DiffSimu, and this can change whether \DDPG or \DiffSimu is best, in Setting~3.  This corroborates our Takeaway~II from the Introduction.

\begin{figure}
\begin{center}

\includegraphics[width=0.75\linewidth]{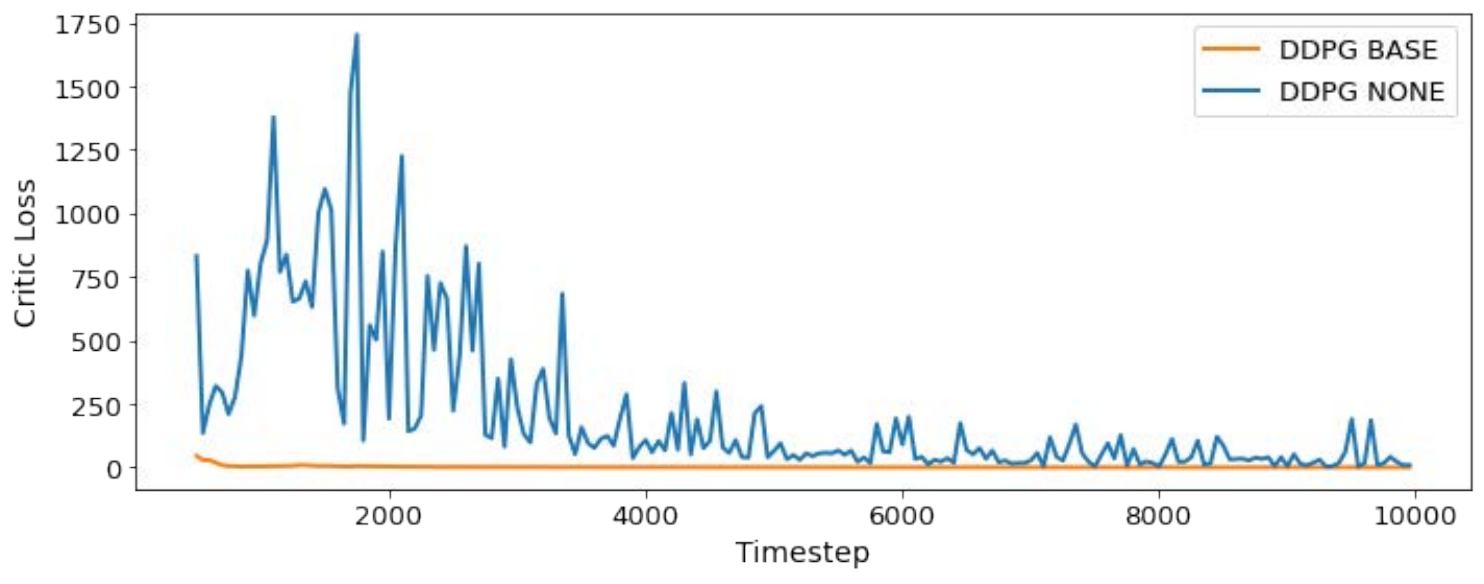}

\end{center}

\begin{center}

\includegraphics[width=0.9\linewidth]{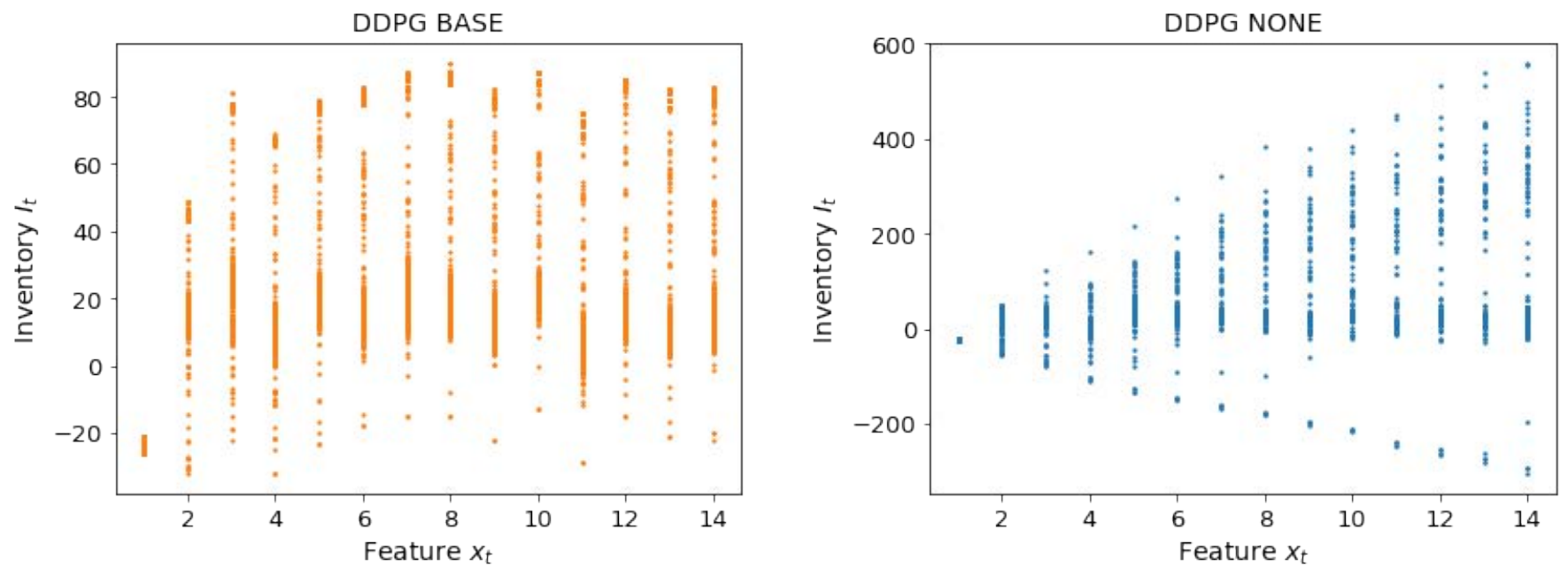}

\end{center}

\caption{Critic loss and states during the first 10,000 timesteps in Setting 1 under a hyperparameter configuration  tuned for \DDPG\None. The training starts after 500 transitions are collected via random actions. Each dot in the two bottom subfigures corresponds to a single visit to state $s=(I_t, x_t)$.}
\label{fig-ddpg-critic-buffer}
\end{figure}

For a more nuanced reason why traditional DRL methods enjoy greater benefit from regularization, we consider \DDPG as an example. \DDPG minimizes the Bellman error of the fitted Q-function over historical transitions $(s, a, r, s')$, where $s$, $a$, and $r$ denote the current state, action, and immediate loss, and $s'$ is the next state. The Bellman error is the gap between $Q(s,a)$ and the target value, defined as the sum of $r$ and the future loss $\min_{a'} Q(s', a')$ (the squared gap is referred to as critic loss). We note that the action $a$ stored in each transition is the output of the policy network $\mu$. That is, with the \Base regularization, the action produced by $\mu_{\Base}$ represents the target level of total inventory, whereas without regularization the action corresponds to the final order quantity.
In \Cref{fig-ddpg-critic-buffer}, we report the critic loss and the states $s=(I_t, x_t)$, where $x_t=\left\lfloor \frac{t-1}{2}\right\rfloor$, stored in the replay buffer over the first 10,000 timesteps (i.e., 10,000 transitions) under a tuned hyperparameter configuration in Setting 1. We observe that the critic loss under \Base regularization is consistently lower, indicating faster convergence of the Q-function. This improvement arises because \Base regularization constrains the inventory state $I_t$ to a narrower range, consistent with the definition of a “Base Stock” policy and confirmed by the bottom subfigures of \Cref{fig-ddpg-critic-buffer}. Consequently, each single state  in the range is visited more frequently, leading to more effective sampling and learning.

Having established the benefit of policy regularization to traditional DRL methods, we now compare their final performance against \DiffSimu across different settings.
In Setting~3 (with the \Base regularization),  we can see that \DiffSimu has better Validation Loss than \DDPG but worse Testing Loss, suggesting that \DiffSimu is overfitting.
This occurs because \DiffSimu takes policy gradients directly with respect to the trajectory losses in $\cD^\train$, forcing in-sample minimization. Moreover, since \DiffSimu does not learn from $(s, a, r, s')$ transitions, it lacks the cross-temporal learning that is central to traditional DRL methods and therefore DS is less sample-efficient. This overfitting seems to subside when the sample size increases from 10 to 20, as \DiffSimu performs best in Setting~2.

\begin{figure}
    \centering
\includegraphics[width=0.7\linewidth]{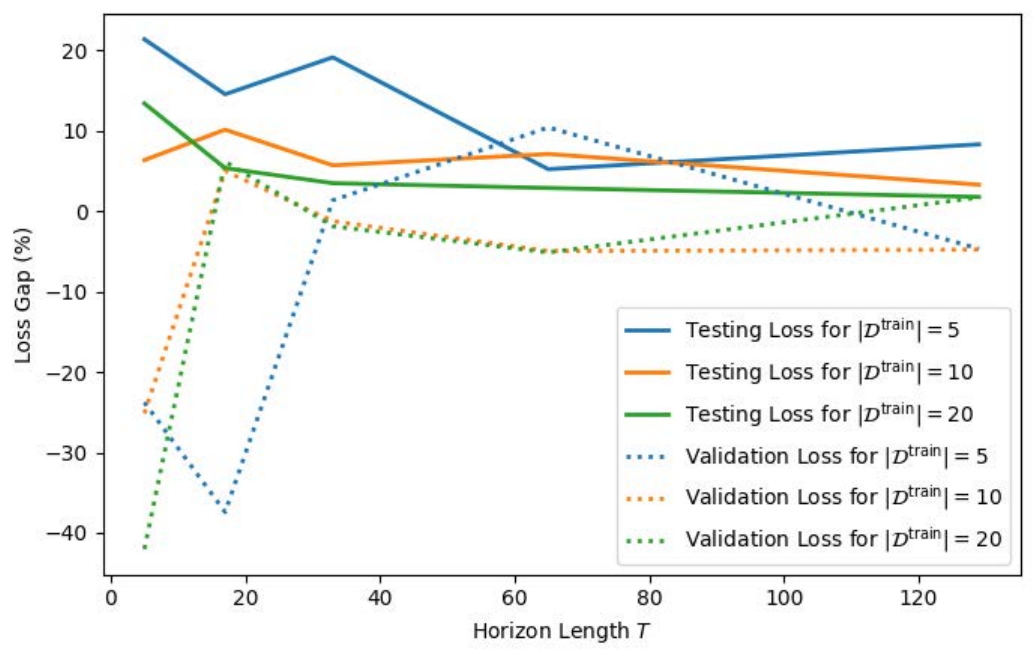}
    \caption{Testing and Validation Loss Gaps for $\DiffSimu$ with $ \Base$ Regularization in Setting 4.}
   \label{fig-exp4-diffNT}
\end{figure}

To further investigate the learning behavior of \DiffSimu, we conclude by training \DiffSimu (with the \Base regularization) under different sample sizes and horizon lengths. In \Cref{fig-exp4-diffNT}, when the sample size is small ($|\cD^\train|=5$), the testing-loss gap remains about 8\% even at $T=129$ under the simple IID distribution.
This shows that increasing the horizon length $T$ yields only limited improvement, highlighting the insufficiency of cross-temporal learning within \DiffSimu. When $|\cD^\train|$ increases from 5 to 10 to 20, the testing losses improve overall across all horizon lengths, indicating that \DiffSimu does benefit from meta-learning across trajectories (i.e., across SKU's), but nonetheless we do not see convergence for longer horizon lengths. As a final remark about overfitting, we find that with a limited number of samples, \DiffSimu exhibits a large gap between validation and testing losses, demonstrating its tendency to overfit to idiosyncrasies in the observed trajectories.

\section{Experiments on Alibaba's Offline Data}

We show one edition of Alibaba's internal testing results on offline data.
Here, $\cD^\train$ consists of 50,000 randomly-selected SKU trajectories from July to September 2024, while
$\cD^\validate$ consists of 5,000 other SKU trajectories from the same horizon of $T=90$ days.
$\cD^\test$ consists of 5,000 randomly-selected SKU trajectories from February to April 2025, which is a chronologically-later horizon.
SKU's in the datasets have varying values for review period $P$ and lead time $L$ that are encoded into their context vectors $x_t$, which lie in $\bR^{190}$.
Typically, review period $P$ takes values of 4 or 7 days, and lead time $L$ takes values of 5 or 7 days.

We compare policies $\pi^{\DRL,\Reg}$ for combinations of $\DRL\in\{\DDPG,\DiffSimu\}$ and $\Reg\in\{\None,\Base,\Coeff,\Both\}$, noting that Alibaba has stopped testing \PPO at scale due to it generally performing worse than \DDPG.
The policies were trained using an objective where a 1\% decrease in Stockout Rate is worth roughly a 2-day decrease in Turnover Time,
and we report the metrics $L^\test_\stockout$ and $L^\test_\turnovertime$ in \Cref{tab:aliOfflineData}.

\begin{table}
\renewcommand{\arraystretch}{0.8}
\begin{tabular}{|c|c|c|c|c|c|c|c|}
\hline
DRL Method & \multicolumn{3}{c|}{\DDPG}                               & \multicolumn{4}{c|}{\DiffSimu}    \\
\hline
Policy Regularization & \None & \Base & \Coeff & \None & \Base & \Coeff & \Both \\ \hline
$L^\test_\stockout(\pi^{\DRL,\Reg})-L^\test_\stockout(\pi^{\DDPG,\Both})$ (\%)    & \textcolor{red}{10.10}   & \textcolor{red}{6.03}    & \textcolor{red}{4.41}
& \textcolor{red}{2.10}  & \textcolor{red}{2.18}   & \textcolor{red}{1.74}       & \textcolor{red}{1.91}       \\ \hline
$L^\test_\turnovertime(\pi^{\DRL,\Reg})-L^\test_\turnovertime(\pi^{\DDPG,\Both})$ (days)  & \textcolor{red}{6.13}   & \textcolor{red}{6.46}    & \textcolor{Green}{-0.41}
& \textcolor{Green}{-1.25}  & \textcolor{Green}{-2.81}   & \textcolor{red}{3.80}       & \textcolor{red}{0.23}       \\ \hline
\end{tabular}
\caption{
Stockout Rates (reported in \% of days) and Turnover Times (reported in days) on the test data, comparing all policies to $\pi^{\DDPG,\Both}$.  Positive values mean worse than $\pi^{\DDPG,\Both}$; negative values mean better.
}
\label{tab:aliOfflineData}
\end{table}
\renewcommand{\arraystretch}{1.0}

\paragraph{Results.} As shown in \Cref{tab:aliOfflineData}, having \Both regularizations produces by far the best version of \DDPG, while having the \Base regularization produces the best version of \DiffSimu.  This supports our Takeaway~I from the Introduction.  To see the result of Takeaway~II, note that without regularizations, $\pi^{\DDPG,\None}$ is much worse than $\pi^{\DiffSimu,\None}$; with regularizations, $\pi^{\DDPG,\Both}$ is comfortably better than $\pi^{\DiffSimu,\Base}$ (and $\pi^{\DiffSimu,\Both}$), recalling that Stockout Rate is twice as important as Turnover Time.

Finally, related to Takeaway~I, one may wonder whether the results in \Cref{tab:aliOfflineData} are from "limited hyperparameter search".
To give some perspective, one hyperparameter trial for \DDPG in this setting (with data size $\approx\text{50,000 trajectories}\times 90\text{ days}\times 190\text{-dimensional features}$) takes about 24 wall-clock hours internally at Alibaba.
\Cref{tab:aliOfflineData} shows test results for every $\pi^{\DRL,\Reg}$
after 10 sequential trials, with each trial choosing new hyperparameters based on the outcome of the previous trial, over 10 physical days.
Although Alibaba could have potentially improved the hyperparameter search with some parallelization, which may decrease the improvement of $\pi^{\DDPG,\Both}$ over $\pi^{\DDPG,\None}$, this would only increase the advantage of \DDPG over \DiffSimu.
Regardless, training is expensive and slow at this scale, and the results in \Cref{tab:aliOfflineData} depict a realistic outcome for hyperparameter tuning in practice at Alibaba.

\section{Real-world Deployment and Impact at Alibaba} \label{sec:layer4}
We report deployment results observed in the real world. Unlike the offline evaluations presented in \Cref{tab:aliOfflineData}, in actual deployment we may have non-deterministic review periods and lead times.
Another difference is that we encounter true demands in the real world, allowing us to stress-test the robustness of our offline training using inferred demands when there was a stockout.
Finally, we note that the horizon length $T$ used for training can be arbitrary, where each trained policy is executed in the real world until the next retraining.
Alibaba aims to retrain every 1.5 months, using data from the most recent 90 days.

To get a sense of real-world scale, Alibaba takes full ownership of inventory and sales for over 100,000 SKU's on its Tmall e-commerce platform.  Inventory for these SKU's is controlled separately at each of Tmall's $\approx 20$ primary warehouses, leading to over 1 million SKU-warehouse pairs that require periodic inventory decisions.

We distinguish between SKU's sourced from domestic suppliers vs.\ SKU's imported from international suppliers, because the incumbent inventory policies for domestic vs.\ international SKU's were different.
The incumbent inventory policies deployed by Alibaba at that time consisted of either DRL-based algorithms or the more traditional predict-then-optimize approach, in which a demand model is estimated from recent demand sequences and the optimal order quantity for each SKU is subsequently computed using standard optimization tools.
The same predict-then-optimize approach can also differ in the depth of its search to manage computational cost. The incumbent international policies adopted more sophisticated search procedures that are computationally intensive, whereas the domestic counterparts used more naive but efficient search schemes to accommodate the much higher volume of domestic products.
To better position our proposed method, we next briefly review Alibaba’s prior DRL implementations and explain why the new policy-regularized RL framework is substantially more scalable.

\paragraph{History of DRL Implementations at Alibaba.}
Alibaba has been implementing RL methods on a subset of its product portfolio since 2020, as documented in \cite{liu2023ai}. The earlier version of the RL algorithm was not general-purpose, requiring similar products to be grouped first before training separate models for each group. This grouping was necessary because different products exhibit significant variations in business objectives, demand patterns, and supply constraints. For example, in terms of business objectives, bestsellers prioritize stock availability, whereas long-tail products emphasize inventory turnover;
on the supply side, fulfillment capabilities vary across merchants, with substantial differences in restocking frequency, delivery times, and minimum order quantities. Moreover, cross-group transfer from bestsellers to long-tail products is challenging, because the long-tail group suffers from data sparsity and poor generalization. These challenges necessitate maintaining independent feature pipelines, actor/critic networks, and hyperparameter tuning processes for each group, which significantly increases engineering overhead and limits the scalability of RL across all products. Therefore, a significant portion of the portfolio at the time relied on a more traditional predict-then-optimize approach, which was substantially easier to maintain. While an RL model only required retraining every one to two months, each retraining cycle was time-consuming especially in a highly dynamic environment where feature updates were frequent. In contrast, predict-then-optimize models were more lightweight operationally.
To illustrate the difference in maintenance burden: RL deployment required a dedicated team of algorithm scientists, whereas predict-then-optimize models could be maintained by a team of engineers plus one or two scientists.

Compared to the earlier version, the new version of RL (as discussed in this paper) is general-purpose: a single model applies across all products, eliminating the need to train separate models. This is enabled by the idea of policy regularizations, supported by several engineering components, including (1) integrating with the operational system so that business objectives (e.g., service levels, inventory turnover) set on the system side are automatically reflected as reward weights, (2) enriching state features to capture product, supplier, and sales channel diversity, (3) scaling to larger-scale deep networks (up to billions of parameters) for robust high-dimensional mappings, and (4) shifting from manual hyperparameter tuning to automated Taguchi experiments.

\paragraph{Results 1: Difference-in-Differences (DiD) study in July 2024.}
An inventory policy trained through DRL with our policy regularizations was piloted on 10\% of international SKU's, cherry-picked to be problematic with great room for inventory improvement.
We compare this 10\% of SKU's to the other 90\% of international SKU's before and after the deployment of our policy, and a simple DiD analysis reveals that the piloted SKU's saw a 0.83\% reduction in average Stockout Rate and a 9.53-day reduction in average Turnover Time.
We note that this is a significant, Pareto improvement in both of our metrics.

\paragraph{Results 2: major rollout in April 2025.}
An inventory policy trained through our DRL was rolled out to 100\% of international SKU's and 87\% of domestic SKU's.
Although this major rollout impedes comparison to the incumbent algorithm via DiD, Alibaba conducted a careful counterfactual analysis that simulated the incumbent algorithm with the non-deterministic review periods and lead times actually encountered during April 2025.  %

We compare the actual metrics observed by our deployed policy to this counterfactual execution of the incumbent algorithm during April 2025.
Averaging over international SKU's, our policy achieved a 1-day reduction in Turnover Time compared to the incumbent algorithm,
with no distinguishable difference in Stockout Rate.
Averaging over domestic SKU's, our policy achieved a 2-day reduction in Turnover Time, also with no distinguishable difference in Stockout Rate.
A further breakdown by sales volume is shown in \Cref{tab:turnover_onshelf_role}, where we note that these simple averages undersell our improvement in Turnover Time because the improvement is greatest on the A+ (highest-volume) SKU's.
We also note that the improvement in domestic SKU's is greater partly because the incumbent algorithm is more naive, even though the improvement is not to the level of the international SKU's from the July 2024 DiD study that were cherry-picked.

\renewcommand{\arraystretch}{0.8}
\begin{table}
\centering
\begin{tabular}{|l|c|c|c|c|c|c|}
\hline
\text{SKU Category} & \text{A+} & \text{A} & \text{B} & \text{C} & \text{D} & \text{Z} \\
\hline
\text{International SKU's (change in days)} & \textcolor{Green}{-1.36} & \textcolor{Green}{-0.82} & \textcolor{Green}{-1.02} & \textcolor{Green}{-1.27} & -- & -- \\ \hline
\text{Domestic SKU's (change in days)} & \textcolor{Green}{-4.04} & \textcolor{Green}{-3.81} & \textcolor{Green}{-2.93} & \textcolor{Green}{-2.23} & \textcolor{Green}{-2.13} & \textcolor{Green}{-0.64} \\
\hline
\end{tabular}
\caption{Changes in average Turnover Time without changing average Stockout Rate. SKU's are categorized by sales volume from A+ (fastest-moving) to Z (long-tail). The time period is April 2025.}
\label{tab:turnover_onshelf_role}
\end{table}
\renewcommand{\arraystretch}{1.0} %

This Pareto improvement of reducing Turnover Time without changing Stockout Rate also allows to estimate the financial benefit from the inventory reduction. Assuming an average inventory value of 5 billion RMB, Alibaba estimates an inventory reduction worth 350 million RMB, leading to annual savings in the cost of capital that amounts to approximately 13.3 million RMB.

\begin{figure}
    \centering
\includegraphics[width=0.6\linewidth]{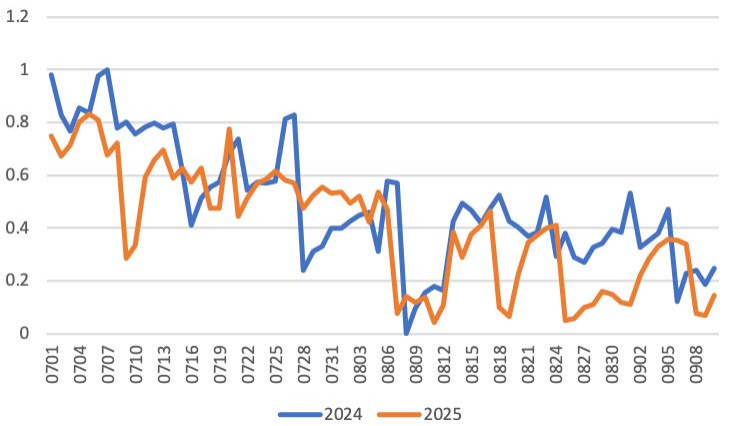}
    \caption{Evolution of average Turnover Time for international SKU's during July--August, in 2024 and 2025.  All numbers are normalized relative to the maximum average Turnover Time encountered in either year.}
    \label{fig:new}
\end{figure}

\paragraph{Results 3: turnover during July--August 2025.}
Alibaba continued monitoring inventory performance after our major rollout of DRL, especially when the inventory policy has to be retrained.
To this end, we show the average Turnover Time of international SKU's during July--August 2025, a period that captures retrainings of the DRL policy, and compare it to the average Turnover Time of international SKU's during the same period in 2024, before our major rollout of DRL.
As seen from \Cref{fig:new}, there is a massive 20\% reduction in Turnover Time in 2025;
it was also reported to us that Stockout Rates did not get noticeably worse.
Although this comparison does not rigorously control for many other factors\footnote{We note that July and August are relatively calm months for Tmall without major promotional events that could have differential effects between 2024 and 2025, which is why this period was chosen for comparison.} that could have caused the change from 2024 to 2025, it provides high-level managerial confidence that our deployment of DRL is stable over time, even when the policy is retrained.  As a result, our policy has been fully deployed as of October 2025 to 100\% of the SKU-warehouse pairs (over 1 million) managed by Alibaba.

\section{Conclusion}
All in all, our paper demonstrates the power of a hybrid approach that integrates classical inventory heuristics from Operations Research to regularize policies learned by powerful modern DRL methods.
Performance aside, we believe that these policy regularizations are essential for operationalizing DRL in practice, as they greatly reduce the computational burden of training and hyperparameter tuning, factors we have observed to be dealbreakers within companies.
Our paper also presents a more detailed comparison of DRL algorithms with and without policy regularizations on synthetic data.
Implementing a regularized DRL algorithm at Alibaba has had a tremendous impact, allowing inventory replenishment decisions to automatically respond to dynamic, high-dimensional demand signals while remaining interpretable and consistently achieving strong performance.

\bibliographystyle{abbrvnat}
\bibliography{mybib}

\newpage

\section{Appendix A: Synthetic Experiments}

\subsection{Implementation and Hyperparameter Details}
Our \DDPG and \PPO code is based on the open-source Stable-Baselines3 implementation \citep{raffin2021stable}, and our \DiffSimu code is based on the open-source implementation of \citet{alvo2023neural}. Both are implemented in PyTorch. The synthetic experiments were conducted on a rented Runpod instance equipped with one NVIDIA RTX A4500 GPU, 12 vCPU's (AMD EPYC 7352), and 62 GB of RAM.

During training, \DDPG and \PPO are evaluated on $\cD^\validate$ every \textsf{eval\_freq} steps. Training is stopped if the validation loss $L_{\bh}^{\validate}$ does not improve over the last  \textsf{patience} evaluations or if the total number of training steps exceeds  \textsf{total\_steps}; in either case, the most recent policy is saved. For \DiffSimu, evaluation on $\cD^\validate$ is performed once per epoch, and training is stopped if $L_{\bh}^{\validate}$ does not improve over the last \textsf{patience} evaluations or if the total number of training epochs exceeds  \textsf{epochs}.

All algorithms are trained using the Adam optimizer with an initial learning rate  \textsf{learning\_rate}. To improve convergence, we adopt a learning-rate scheduler for each algorithm. For \DDPG and \PPO, the scheduler follows the Stable-Baselines3 function \textit{get\_linear\_fn}\textsf{(start, end, end\_fraction)} with $\textsf{start} = \textsf{learning\_rate}$, $\textsf{end} = \textsf{lr\_min}$, and $\textsf{end\_fraction} = \textsf{lr\_fraction}$.
We also schedule the clip range for \PPO using \textit{get\_linear\_fn}\textsf{(start, end, end\_fraction)}, where $\textsf{start} = \textsf{clip\_range}$, $\textsf{end} = 0.5\times \textsf{clip\_range}$, and $\textsf{end\_fraction} = \textsf{lr\_fraction}$.
For \DiffSimu, we use the PyTorch \textit{ReduceLROnPlateau} scheduler with $\textsf{patience} = 3$ and $\textsf{factor} = \textsf{scheduler\_factor}$, keeping all other inputs at their default values.

For consistency, we use the following procedure to tune hyperparameters in all experiments on synthetic data.
We use the Ray Tune package \citep{liaw2018tune} to intelligently search over a hyperparameter grid (see details in \Cref{hyperpara-ddpg,hyperpara-ppo,hyperpara_sl}) and identify the configuration that minimizes validation loss.
The search uses Bayesian optimization with a Tree-structured Parzen Estimator, evaluating 50 configurations in the first stage.
In the second stage, the five best configurations from stage one are each re-run across five random seeds, and the final winner is selected based on median performance across these seeds, using the stage-one configuration with the lowest average validation loss. This two-stage procedure helps mitigate overfitting and ensures robust hyperparameter selection for evaluation.

We specify the hyperparameter values or search ranges used in each setting in \Cref{hyperpara-ddpg,hyperpara-ppo,hyperpara_sl}, where \textit{tune.choice}, \textit{tune.uniform}, and \textit{tune.loguniform} are Ray Tune sampling functions that draw uniformly from a discrete set, a continuous interval, and a continuous interval on the log scale, respectively. We use bold text to indicate hyperparameters that are built-in inputs to the Stable-Baselines3 implementations of \DDPG and \PPO, while all remaining built-in hyperparameters follow their default values (see \url{https://stable-baselines3.readthedocs.io/en/master/}).
In addition to \textsf{eval\_freq},  \textsf{patience}, \textsf{total\_steps}, \textsf{epochs}, \textsf{learning\_rate}, \textsf{lr\_min}, \textsf{lr\_fraction}, and \textsf{scheduler\_factor}, our self-defined hyperparameters also include \textsf{max\_replenish} and  \textsf{initial\_action\_bias}. They ensure that both the action space and the final order quantity are upper bounded by \textsf{max\_replenish}, and that the initial bias of the policy network equals $\textsf{initial\_action\_bias} \times \textsf{max\_replenish}$.

\begin{table}[h]
\centering
\small
\begin{tabular}{|c|c>{\centering\arraybackslash}p{5.5cm}|c|}
\hline
  &  & INDEP & AR(1)\\
\hline
\textbf{\textsf{policy}}         &  & \multicolumn{2}{c|}{`MlpPolicy'} \\  \hline

\multirow{3}{*}{\makecell{\textbf{\textsf{learning\_rate}} =\\ \textit{get\_linear\_fn}}}
    & \multicolumn{1}{c|}{\textsf{learning\_rate}}
    & \textit{tune.loguniform}(3e-4, 9e-3) & \textit{tune.loguniform}(6e-4, 9e-3) \\
    \cline{2-4}
    & \multicolumn{1}{c|}{\textsf{lr\_min}}
    & \textit{tune.loguniform}(6e-5, 3e-4) & \textit{tune.loguniform}(9e-5, 6e-4) \\
    \cline{2-4}
    & \multicolumn{1}{c|}{\textsf{lr\_fraction}}
    & \multicolumn{2}{c|}{0.95} \\  \hline

\textbf{\textsf{batch\_size}}
    &  & \multicolumn{2}{c|}{\textit{tune.choice}([128, 256, 512])}  \\  \hline

\textbf{\textsf{tau}}
    &  & \multicolumn{2}{c|}{\textit{tune.loguniform}(1e-3, 1e-2)} \\ \hline

\textbf{\textsf{gamma}}
    &  & \textit{tune.uniform}(0.98, 1) & \textit{tune.uniform}(0.96, 1) \\  \hline

{\makecell{\textbf{\textsf{policy\_kwargs}} =\\ {dict}[\textsf{net\_arch}]}}
    &  & `pi': [64, 64], `qf': [64, 64]
      & \makecell[c]{
                `pi': [32, 32], `qf': [32, 32]
                for Setting~2\\
                `pi': [32, 32], `qf': [16, 16]
                for Setting~3}
\\ \hline

\textbf{\textsf{buffer\_size}}
    &  & \multicolumn{2}{c|}{\textit{tune.choice}([20,000, 50,000, 100,000])} \\ \hline

\textbf{\textsf{train\_freq}}
    &  & \multicolumn{2}{c|}{\textit{tune.choice}([1, 14])} \\ \hline

\textbf{\textsf{learning\_starts}}
    &  & \multicolumn{2}{c|}{\textit{tune.choice}([100, 500, 1000])} \\ \hline

\textsf{total\_steps}
    &  & \multicolumn{2}{c|}{200,000} \\ \hline

\textsf{eval\_freq}
    &  & \multicolumn{2}{c|}{\textit{tune.choice}([512, 1024])} \\  \hline

\textsf{patience}
    &  & \multicolumn{2}{c|}{10} \\  \hline

\textsf{max\_replenish}
    &  & \multicolumn{2}{c|}{100} \\  \hline

\textsf{initial\_action\_bias}
&  & \textit{tune.uniform}(0.2, 0.7) & \textit{tune.uniform}(0, 0.6)\\ \hline
\end{tabular}
\caption{Hyperparameter values or ranges of \DDPG.}
\label{hyperpara-ddpg}
\end{table}

\begin{table}[h]
\centering
\small
\begin{tabular}{|c|c>{\centering\arraybackslash}p{5.5cm}|c|}
\hline
  &  & INDEP & AR(1)\\
\hline
\textbf{\textsf{policy}}         &  & \multicolumn{2}{c|}{`MlpPolicy'} \\  \hline

\multirow{3}{*}{\makecell{\textbf{\textsf{learning\_rate}} =\\ \textit{get\_linear\_fn}}}
    & \multicolumn{1}{c|}{\textsf{learning\_rate}}
    & \textit{tune.loguniform}(5e-5, 9e-3) & \textit{tune.loguniform}(6e-4, 1e-2) \\
    \cline{2-4}
    & \multicolumn{1}{c|}{\textsf{lr\_min}}
    & \textit{tune.loguniform}(1e-5, 5e-5) & \textit{tune.loguniform}(9e-5, 6e-4) \\
    \cline{2-4}
    & \multicolumn{1}{c|}{\textsf{lr\_fraction}}
    & \multicolumn{2}{c|}{0.9} \\  \hline

\multirow{2}{*}{\makecell{\textbf{\textsf{clip\_range}} =\\ \textit{get\_linear\_fn}}}
    & \multicolumn{1}{c|}{\textsf{clip\_range}}
    &  \multicolumn{2}{c|}{\textit{tune.uniform}(0.1, 0.3)} \\  \cline{2-4}
    & \multicolumn{1}{c|}{\textsf{lr\_fraction}}
    & \multicolumn{2}{c|}{0.9} \\  \hline

\textbf{\textsf{batch\_size}}
    &  & \multicolumn{2}{c|}{\textit{tune.choice}([128, 256, 512])}  \\  \hline

\textbf{\textsf{n\_steps}}
    &  & \textit{tune.choice}([1024, 2048, 4096])  &\textit{tune.choice}([512, 1024, 2048]) \\ \hline

\textbf{\textsf{ent\_coef}}
    &  & \multicolumn{2}{c|}{\textit{tune.uniform}(1e-4, 5e-1)} \\ \hline

\textbf{\textsf{gamma}}
    &  & \textit{tune.uniform}(0.98, 1) & \textit{tune.uniform}(0.96, 1) \\  \hline

\textbf{\textsf{gae\_lambda}}
    &  & \textit{tune.uniform}(0.5, 1) &
    \makecell[c]{ \textit{tune.uniform}(0, 0.5)
                for \Base\\
                \textit{tune.uniform}(0.5, 1)
                for \None} \\  \hline

{\makecell{\textbf{\textsf{policy\_kwargs}} =\\ {dict}[\textsf{net\_arch}]}}
    &  & `pi': [256, 256], `vf': [256, 256]
      & \makecell[c]{
                `pi': [128, 128], `vf': [64, 64]
                for Setting~2\\
                `pi': [64, 64], `vf': [64, 64]
                for Setting~3}
\\ \hline

\textbf{\textsf{vf\_coef}}
    &  & \multicolumn{2}{c|}{0.4} \\ \hline

\textsf{total\_steps}
    &  & \multicolumn{2}{c|}{200,000} \\ \hline

\textsf{eval\_freq}
    &  & \textit{tune.choice}([512, 1024, 2048]) & \textit{tune.choice}([1024, 2048])  \\  \hline

\textsf{patience}
    &  & \multicolumn{2}{c|}{10} \\  \hline

\textsf{max\_replenish}
    &  & \multicolumn{2}{c|}{100} \\  \hline

\textsf{initial\_action\_bias}
&  & \textit{tune.uniform}(0.2, 0.7) & \textit{tune.uniform}(0, 0.6)\\ \hline
\end{tabular}
\caption{Hyperparameter values or ranges of \PPO.}
\label{hyperpara-ppo}
\end{table}

\begin{table}[h]
\centering
\small
\begin{tabular}{|c|c|>{\centering\arraybackslash}p{2.6cm}|c|}
\hline
  &   INDEP & AR(1) & IID \\
\hline
\textsf{policy}
    &   \multicolumn{3}{c|}{`MlpPolicy'} \\
\hline

\textsf{neurons\_per\_hidden\_layer}
      & \makecell[r]{
        tune.choice([[32, 32],\\
        \quad [64, 64], \\
        \quad [128, 128]])}
    &  \multicolumn{2}{c|}{\makecell[c]{
        tune.choice([[16, 16],
        [32, 32],
        [64, 64]])}}
     \\
\hline

\textsf{inner\_layer\_activation}
      & \multicolumn{3}{c|}{`elu'} \\
\hline

\textsf{output\_layer\_activation}
      & \multicolumn{3}{c|}{`relu'} \\
\hline

\textsf{batch\_size}
      & \multicolumn{2}{c|}{\textit{tune.choice}([5, 10])} & \textit{tune.choice}([5, 10, 20])\\
\hline

\textsf{learning\_rate}
     & \multicolumn{3}{c|}{\textit{tune.loguniform}(4e-4, 1e-1)} \\
\hline

\textsf{scheduler\_factor}
      & \multicolumn{2}{c|}{\textit{tune.uniform}(0.5, 0.95)} & 0.5 \\
\hline

\textsf{epochs}
      & \multicolumn{3}{c|}{3000} \\
\hline

\textsf{patience}
      & \multicolumn{3}{c|}{30} \\
\hline

\textsf{max\_replenish}
      & \multicolumn{2}{c|}{100}
    & 500\\
\hline

\textsf{initial\_action\_bias}
      & {\textit{tune.uniform}(0.2, 0.7)}
        &  \multicolumn{2}{c|}{\textit{tune.uniform}(0, 0.6)}
       \\
\hline
\end{tabular}
\caption{Hyperparameter values or ranges of \DiffSimu.}
\label{hyperpara_sl}
\end{table}

\subsection{Trajectory Distributions and Details of Synthetic Results}
We describe the three types of trajectory distributions as follows:
\begin{enumerate}
    \item Independent but non-identically distributed demands (INDEP distribution):
    for all $(|\cD^{\train}| + |\cD^{\validate}| + |\cD^{\test}|)$ SKU's, demand trajectories are IID drawn from the same distribution.
    At time $t$, the demand $d_t$ is independently drawn from a discrete uniform distribution $U\{b_t, b_t+1, \dots, b_t+4\}$. Here, the lower bound $b_t \in \mathbb{Z}_{\ge 0}$ is generated from a truncated normal distribution $\left\lfloor \max\{0, \mathcal{N}(\mu,\sigma^2)\}\right\rfloor$ with  $\mu=10$ and  $\sigma=4$.

    \item Autoregressive demands (AR(1) distribution): for each SKU $i\in \left\{1, ..., |\cD^{\train}| + |\cD^{\validate}| + |\cD^{\test}|\right\}$, the demands $(d_t^i)_{t=1}^T$ follow a truncated AR(1) process. Specifically, $
    d^i_t = \max\{0, \phi^i d^i_{t-1} + (1-\phi^i)\, \bar{d}^i + \epsilon^i_t\}$,
   where the initial demand $d_0^i$ is drawn from $\max\left\{0,\mathcal{N}(\bar{d^i}, (\sigma^i)^2)\right\}$.
    Here, the coefficient $\phi^i$, base demand  $\bar{d^i}$, and noise standard deviation $\sigma^i$ are drawn from uniform distributions $U[0.3,0.9]$,   $U[5, 10]$ and  $U[2,8]$, respectively, and $\epsilon_t^i$ follows $\mathcal{N}(0, (\sigma^i)^2)$.

    \item Independent and identically distributed demands (IID distribution): for each SKU $i\in \{1, ..., |\cD^{\train}|\}$, the demand $d^i_t$ at every time $t$ is independently drawn from a truncated normal distribution $  \max\left\{0, \mathcal{N}(100,(\sigma^i)^2)\right\}$, where $\sigma^i$ is drawn from a uniform distribution $U[5, 20]$. Note that the feature $x_t^i$ of SKU $i$ here is set to be the parameter $\sigma^i$ for each $t$.%

\end{enumerate}

The absolute values of the validation and testing losses corresponding to the percentage loss gaps in \Cref{fig-exp1-valtest,fig-exp4-diffNT} are reported in \Cref{table-exp1-valtest,table-exp2-valtest-largedata,table-exp2-valtest-smalldata,table-exp4-DS}. The loss gaps are computed relative to the benchmark $L_{\bh}^{\test} (\pi^*)$ obtained by solving or approximating the optimal policy $\pi^*$ with knowledge of the true demand distribution.
For the INDEP distribution, it is well known that the optimal policy belongs to the class of time-dependent base-stock policies (see, e.g., \citealt{snyder2019fundamentals}), and we compute it exactly via dynamic programming.
For the AR(1) distribution, the optimal loss is approximated by training \DiffSimu\ on the testing dataset $\cD^{\test}$.
For the IID distribution, it is well known that the optimal policy lies within the class of stationary base-stock policies (see, e.g., \citealt{snyder2019fundamentals}), and we compute the optimal base-stock level exactly via the critical fractile formula.

\begin{table}[h]
\centering
\begin{tabular}{|c|cc|cc|cc|}
\hline DRL Method
 & \multicolumn{2}{c|}{\DDPG} & \multicolumn{2}{c|}{\PPO} & \multicolumn{2}{c|}{\DiffSimu} \\ \cline{1-7}
 Policy Regularization                 & \multicolumn{1}{c|}{\None} & \multicolumn{1}{c|}{\Base} & \multicolumn{1}{c|}{\None} & \multicolumn{1}{c|}{\Base} & \multicolumn{1}{c|}{\None} & \multicolumn{1}{c|}{\Base} \\ \hline

Validation Loss       & \multicolumn{1}{c|}{0.9340} & 0.9264 & \multicolumn{1}{c|}{ 0.9045 } & 0.8917 & \multicolumn{1}{c|}{ 1.2371 } & 1.0454 \\ \cline{1-7}

Validation Loss Gap (\%)              & \multicolumn{1}{c|}{7.73} &  6.86  & \multicolumn{1}{c|}{4.33} & 2.85 & \multicolumn{1}{c|}{42.69} & 20.57  \\ \cline{1-7}

Testing Loss             & \multicolumn{1}{c|}{0.9443} & 0.9356 & \multicolumn{1}{c|}{0.9104} & 0.8966 & \multicolumn{1}{c|}{1.2540} & 1.0959\\ \cline{1-7}

Testing Loss Gap (\%)             & \multicolumn{1}{c|}{8.92} & 7.92  & \multicolumn{1}{c|}{5.01} & 3.41 & \multicolumn{1}{c|}{44.64} & 26.40  \\ \hline
\end{tabular}
\caption{Validation and Testing Loss and corresponding Loss Gaps relative to the benchmark loss of 0.8670 in Setting~1.}
\label{table-exp1-valtest}
\end{table}

\begin{table}[h]
\centering
\begin{tabular}{|c|cc|cc|cc|}
\hline  DRL Method
 & \multicolumn{2}{c|}{\DDPG} & \multicolumn{2}{c|}{\PPO} & \multicolumn{2}{c|}{\DiffSimu} \\ \cline{1-7}
  Policy Regularization               & \multicolumn{1}{c|}{\None} & \multicolumn{1}{c|}{\Base} & \multicolumn{1}{c|}{\None} & \multicolumn{1}{c|}{\Base} & \multicolumn{1}{c|}{\None} & \multicolumn{1}{c|}{\Base} \\ \hline

Validation Loss      & \multicolumn{1}{c|}{2.3139 } & 2.2647 & \multicolumn{1}{c|}{2.4428} & 2.3069 & \multicolumn{1}{c|}{2.2561} & 2.2550 \\ \cline{1-7}

Validation Loss Gap  (\%)            & \multicolumn{1}{c|}{1.32} & -0.83 & \multicolumn{1}{c|}{6.96} & 1.01 & \multicolumn{1}{c|}{-1.21} & -1.26 \\ \cline{1-7}

Testing Loss               & \multicolumn{1}{c|}{2.5735} & 2.4557 & \multicolumn{1}{c|}{2.7257} & 2.5273 & \multicolumn{1}{c|}{2.4455} & 2.4342 \\ \cline{1-7}

Testing Loss Gap  (\%)           & \multicolumn{1}{c|}{12.69} & 7.53 & \multicolumn{1}{c|}{19.35} & 10.66 & \multicolumn{1}{c|}{7.08} & 6.58 \\ \hline
\end{tabular}
\caption{Validation and Testing Loss and corresponding Loss Gaps relative to the benchmark loss of 2.2838 in Setting~2.}
 \label{table-exp2-valtest-largedata}
\end{table}

\begin{table}[h]
\centering
\begin{tabular}{|c|cc|cc|cc|}
\hline  DRL Method
 & \multicolumn{2}{c|}{\DDPG} & \multicolumn{2}{c|}{\PPO} & \multicolumn{2}{c|}{\DiffSimu} \\ \cline{1-7}
  Policy Regularization               & \multicolumn{1}{c|}{\None} & \multicolumn{1}{c|}{\Base} & \multicolumn{1}{c|}{\None} & \multicolumn{1}{c|}{\Base} & \multicolumn{1}{c|}{\None} & \multicolumn{1}{c|}{\Base} \\ \hline

Validation Loss      & \multicolumn{1}{c|}{2.2341} & 2.3236 & \multicolumn{1}{c|}{2.3515} & 2.4747 & \multicolumn{1}{c|}{2.1643} & 2.2141 \\ \cline{1-7}

Validation Loss Gap  (\%)            & \multicolumn{1}{c|}{-2.18} & 1.74  & \multicolumn{1}{c|}{2.97} & 8.36 & \multicolumn{1}{c|}{-5.23} & -3.05 \\ \cline{1-7}

Testing Loss               & \multicolumn{1}{c|}{2.6275} & 2.4727 & \multicolumn{1}{c|}{2.6664} & 2.5319 & \multicolumn{1}{c|}{2.5017} & 2.5060 \\ \cline{1-7}

Testing Loss Gap  (\%)           & \multicolumn{1}{c|}{15.05} & 8.27 & \multicolumn{1}{c|}{16.75} & 10.86 & \multicolumn{1}{c|}{9.54} & 9.73 \\ \hline
\end{tabular}
\caption{Validation and Testing Loss and corresponding Loss Gaps relative to the benchmark loss of 2.2838 in Setting~3.}
\label{table-exp2-valtest-smalldata}
\end{table}

\begin{table}[h]
\centering
\begin{tabular}{|cc|c|c|c|c|c|}
\hline
\multicolumn{2}{|c|}{}                                 & $T=5$ & $T=17$ & $T=33$ & $T=65$ & $T=129$ \\ \hline

\multicolumn{1}{|c|}{\multirow{5}{*}{$|\cD^{\train}|=5$}} & Benchmark Loss  & 2.3669   & 2.3726    & 2.3792    & 2.3884   & 2.3764   \\ \cline{2-7}

\multicolumn{1}{|c|}{}                           & Validation Loss    & 1.8045   & 1.4850    & 2.4117    & 2.6374   & 2.2657     \\ \cline{2-7}

\multicolumn{1}{|c|}{}                           & Validation Loss Gap  (\%)    & -23.76   & -37.41    & 1.37    & 10.43    & -4.66     \\ \cline{2-7}

\multicolumn{1}{|c|}{}                           & Testing Loss Gap   & 2.8733   & 2.7180    & 2.8351    & 2.5135    & 2.5745     \\ \cline{2-7}

\multicolumn{1}{|c|}{}                           & Testing Loss Gap  (\%)   & 21.40   & 14.56    & 19.16    & 5.24    & 8.33    \\ \hline
\multicolumn{1}{|c|}{\multirow{5}{*}{$|\cD^{\train}|=10$}} & Benchmark Loss   & 2.2569   & 2.2672    & 2.2712    & 2.2728    & 2.2640     \\ \cline{2-7}

\multicolumn{1}{|c|}{}                           & Validation Loss    & 1.6897   & 2.3809    & 2.2431    & 2.1605    & 2.1559     \\ \cline{2-7}

\multicolumn{1}{|c|}{}                           & Validation Loss Gap  (\%)    & -25.13   & 5.02    & -1.23    & -4.94    & -4.77     \\ \cline{2-7}

\multicolumn{1}{|c|}{}                           & Testing Loss   & 2.4006   & 2.4975    & 2.4011    & 2.4349    & 2.3389     \\ \cline{2-7}

\multicolumn{1}{|c|}{}                           & Testing Loss Gap  (\%)   & 6.37   & 10.16    & 5.72    & 7.13    & 3.31     \\ \hline
\multicolumn{1}{|c|}{\multirow{5}{*}{$|\cD^{\train}|=20$}} & Benchmark Loss   & 2.0932   & 2.1064    & 2.1011    & 2.0999    & 2.1003     \\ \cline{2-7}

\multicolumn{1}{|c|}{}                           & Validation Loss   & 1.2135  & 2.2363   & 2.0620   & 1.9926   & 2.1363  \\ \cline{2-7}

\multicolumn{1}{|c|}{}                           & Validation Loss Gap  (\%)    & -42.02   & 6.17   & -1.86    & -5.11    & 1.72     \\ \cline{2-7}

\multicolumn{1}{|c|}{}                           & Testing Loss   & 2.3741   & 2.2194    & 2.1747    & 2.1607    & 2.1378    \\ \cline{2-7}

\multicolumn{1}{|c|}{}                           & Testing Loss Gap  (\%) & 13.42 & 5.37  & 3.50  & 2.90  & 1.79   \\ \hline
\end{tabular}
\caption{Validation and Testing Loss and corresponding Loss Gaps for \DiffSimu with \Base regularization in Setting~4.}
\label{table-exp4-DS}
\end{table}

\end{document}